\documentclass[twocolumn]{ceurart}

\usepackage{listings}
\usepackage[utf8]{inputenc} 
\usepackage[T1]{fontenc}    
\usepackage{hyperref}       
\usepackage{url}            
\usepackage{booktabs}       
\usepackage{amsfonts}       
\usepackage{nicefrac}       
\usepackage{microtype}      
\usepackage{xcolor}         
\usepackage{xspace}         
\usepackage{enumitem}       
\usepackage{amsmath}
\usepackage{amssymb}
\usepackage{multirow}
\usepackage{graphicx}
\usepackage[nohyperlinks]{acronym}
\makeatletter
\DeclareRobustCommand\onedot{\futurelet\@let@token\@onedot}
\def\@onedot{\ifx\@let@token.\else.\null\fi\xspace}

\def\eg{\emph{e.g}\onedot} 
\def\ie{\emph{i.e}\onedot}

\def\etal{\emph{et al}\onedot}
\makeatother

\DeclareMathOperator*{\argmax}{arg \ max}
\acrodef{tsr}[TSR]{table structure recognition}
\acrodef{teds}[TEDS]{tree-edit-distance-based similarity}
\acrodef{vlp}[VLP]{vision-language pretraining}
\acrodef{cnn}[CNN]{convolutional neural network}
\acrodef{pp}[pp]{percentage points}
\acrodef{sota}[SOTA]{state-of-the-art}
\acrodef{vit}[ViT]{vision transformer}
\acrodef{relu}[ReLU]{rectified linear unit}
\acrodef{gelu}[GELU]{gaussian error linear unit}
\acrodef{nlp}[NLP]{natural language processing}
\acrodef{vilt}[ViLT]{vision-and-language transformer}
\acrodef{ssl}[SSL]{self-supervised learning}
\acrodef{rf}[RF]{receptive field}
\acrodef{iou}[IoU]{intersection over union}
\acrodef{mac}[MAC]{Multiply-Add Operations per Second}
\acrodef{edd}[EDD]{encoder-dual-decoder}
\acrodef{pdf}[PDF]{portable document format}
\acrodef{ssp}[SSP]{self-supervised pre-training}
\acrodef{mim}[MIM]{masked image modeling}
\acrodef{vqvae}[VQ-VAE]{Vector Quantized-Variational AutoEncoder}

\begin{document}
\copyrightyear{2024}
\copyrightclause{Copyright for this paper by its authors.
  Use permitted under Creative Commons License Attribution 4.0
  International (CC BY 4.0).}

\conference{The AAAI-24 Workshop on Scientific Document Understanding}

\title{Self-Supervised Pre-Training for Table Structure Recognition Transformer}

\author[1]{ShengYun Peng}[
email=speng65@gatech.edu,
url=shengyun-peng.github.io
]

\author[1]{Seongmin Lee}[
email=seongmin@gatech.edu,
url=https://seongmin.xyz/
]

\author[2]{Xiaojing Wang}[
email=xiaojing.wang@adp.com
]

\author[2]{Rajarajeswari Balasubramaniyan}[
email=raji.balasubramaniyan@adp.com
]

\author[1]{Duen Horng Chau}[
email=polo@gatech.edu,
url=https://poloclub.github.io/polochau/
]

\address[1]{Georgia Institute of Technology}
\address[2]{ADP, Inc.}

\begin{abstract}
Table structure recognition (TSR) aims to convert tabular images into a machine-readable format.
Although hybrid \ac{cnn}-transformer architecture is widely used in existing approaches, linear projection transformer has outperformed the hybrid architecture in numerous vision tasks due to its simplicity and efficiency. 
However, existing research has demonstrated that a direct replacement of \ac{cnn} backbone with linear projection leads to a marked performance drop.
In this work, we resolve the issue by proposing a \ac{ssp} method for TSR transformers. 
We discover that the performance gap between the linear projection transformer and the hybrid \ac{cnn}-transformer can be mitigated by \ac{ssp} of the visual encoder in the TSR model. 
We conducted reproducible ablation studies and open-sourced our code at \url{https://github.com/poloclub/unitable} 
to enhance transparency, inspire innovations, and facilitate fair comparisons in our domain as tables are a promising modality for representation learning.
\end{abstract}

\maketitle

\section{Introduction}
\label{sec:intro}

\Acf{tsr} aims to extract the structure of a tabular image into a machine-readable format~\cite{zhong2020image, nassar2022tableformer, huang2023improving}. 
This task is inherently an image-to-text generation problem, where a visual encoder extracts image features, and a textual decoder generates tokens representing the table~\cite{li2019tablebank, ye2021pingan}.
In the existing literature~\cite{nassar2022tableformer, huang2023improving}, the visual encoder is a stack of transformer encoder layers, preceded by a \ac{cnn} backbone, namely the hybrid \ac{cnn}-transformer architecture, and the textural decoder consists of a stack of transformer decoders~\cite{vaswani2017attention}.
However, the linear projection has supplanted the role of \ac{cnn} in this hybrid CNN-transformer after \ac{vit}~\cite{dosovitskiy2020image} came out, owing to its simplicity and effectiveness in multiple vision tasks~\cite{li2021align, bao2021beit, wang2023image}.
Recent work in \ac{tsr}~\cite{peng2023high} has compared these two types of architectures and revealed that the linear projection transformer exhibits a noticeable drop in performance compared to the hybrid \ac{cnn}-transformer. 
We hypothesize this is due to transformers being data-hungry as large \ac{vit} models perform less optimally than ResNet variants when trained on small datasets but outshine them when trained on larger datasets~\cite{dosovitskiy2020image}. 
Considering that the scale of annotated table datasets (PubTabNet~\cite{zhong2020image} with 0.5M and SynTabNet~\cite{nassar2022tableformer} with 0.6M) significantly lags behind those used for training large transformers (ImageNet~\cite{deng2009imagenet} with 1.3M, JFT~\cite{sun2017revisiting} with 300M, and LAION~\cite{schuhmann2022laion} with 5B), a crucial question arises: 
How can we bridge the performance gap without acquiring additional data annotations?

We address the above research question and make the following major contributions:
\begin{enumerate}[leftmargin=*,topsep=0pt]
\itemsep-0.1em 
\item \textbf{\Acf{ssp} for \ac{tsr} transformer.}
We discover that the performance gap between the linear projection transformer and the \ac{cnn}-transformer can be mitigated by \ac{ssp} of the visual encoder. 
The visual encoder is only pre-trained on the tabular image of PubTabNet and SynthTabNet, thus no additional table annotations are required. 
Compared with the same model that is trained from scratch, the performance of the \ac{ssp} model is 12.50 higher in complex \ac{teds} score and 9.76 higher in total \ac{teds} score, which demonstrates the effectiveness of \ac{ssp} in training \ac{tsr} models. 

\item \textbf{Reproducible research and open-source code.}
We provide all the details regarding training, validation, and testing, which include model architecture configurations, model complexities, dataset information, evaluation metrics, training optimizer, learning rate, and ablation studies. 
Our work is open source and publicly available at \url{https://github.com/poloclub/unitable}.
We believe that reproducible research and open-source code enhance transparency, inspire \ac{sota} innovations, and 
facilitate fair comparisons in our domain as tables are a promising modality for representation learning.

\end{enumerate}

\section{Related Work}
\label{related_work}

\ac{tsr}, as an image-to-text generation task, treats the table structure as a sequence and adopts an end-to-end image-to-text paradigm. 
Deng, \etal~\cite{deng2019challenges} employed a hybrid CNN-LSTM architecture to generate the LaTeX code of the table. 
Zhong, \etal~\cite{zhong2020image} introduced an \ac{edd} architecture in which two RNN-based decoders were responsible for logical and cell content, respectively. 
Both TableFormer~\cite{nassar2022tableformer} and TableMaster~\cite{ye2021pingan} enhanced the \ac{edd} decoder with a transformer decoder and included a regression decoder to predict the bounding box instead of the content.
VAST~\cite{huang2023improving} took a different approach by modeling the bounding box coordinates as a language sequence and proposed an auxiliary visual alignment loss to ensure that the logical representation of the non-empty cells contains more local visual details.
Peng, \etal~\cite{peng2023high} designed a lightweight visual encoder without sacrificing the expressive power of hybrid \ac{cnn}-transformer architecture, and the performance of a convolutional stem can match classic CNN backbone performance, with a much simpler model.
Our goal differs from existing approaches as we aim to extend the success of linear projection transformer to \ac{tsr} domain. 
However, a direct application of a linear projection transformer leads to a significant performance drop.
We mitigate the performance gap by leveraging \ac{ssp} for \ac{tsr} transformer. 

\begin{table*}[!htbp]
\small
\centering
\caption{
The \ac{ssp} method (bottom two rows) successfully mitigates the performance gap between the linear projection transformer and the hybrid \ac{cnn}-transformer.
Training the linear projection transformer from scratch leads to a significant performance drop in \ac{teds} score compared to the hybrid \ac{cnn}-transformer architecture, especially for complex tables. 
We present the results of two finetuning methods: 1) LinearProj (frozen) - freezing the weights in the visual encoder and only training the textual decoder for a few epochs, and 2) LinearProj - directly finetuning all the weights.
Both methods achieve similar performance, which outperforms TableFormer and is on par with VAST. 
}
\begin{tabular}{lrrrr}
\toprule
    & & \multicolumn{3}{c}{\acs{teds} (\%)} \\
    Model & \#Param. & Simple & Complex & All \\
\midrule
    EDD~\cite{zhong2020image} & Not reported & 91.1 & 88.7 & 89.90 \\
    GTE~\cite{zheng2021global} & Not reported & - & - & 93.01 \\
    Davar-Lab~\cite{jimeno2021icdar} & Not reported & 97.88 & 94.78 & 96.36 \\
    TableFormer~\cite{nassar2022tableformer} & Not reported & 98.50 & 95.00 & 96.75 \\
    VAST~\cite{huang2023improving}  & Not reported & - & - & 97.23 \\
    CNN-transformer (ResNet-18)~\cite{peng2023high} & 29M & 98.31 & 94.50 & 96.45 \\
\midrule
    LinearProj (from scratch) & 124M & 91.33 & 82.61 & 87.07 \\
    LinearProj (frozen) & 124M & 98.48 & 95.11 & 96.83 \\
    LinearProj & 124M & 98.52 & 95.02 & 96.81 \\
\bottomrule
\end{tabular}
\label{tab:sota}
\end{table*}
\section{TSR Transformer}
\label{sec:method}

\subsection{Model Architecture}
The goal of \ac{tsr} is to translate the input tabular image $I$ into a machine-readable sequence $T$. 
Specifically, $T$ includes table structure $T_s$ defined by HTML table tags and table content $T_c$ defined by standard \ac{nlp} vocabulary. 
The prediction of $T_c$ is triggered when a non-empty table cell is encountered in $T_s$, \eg, \texttt{<td>} for a single cell or \texttt{>} for a spanning cell. 
Thus, accurately predicting $T_s$ is a bottleneck that affects the performance of the prediction of $T_c$. 
Our model focuses on $T_s$ and comprises two modules: visual encoder and textual decoder. 
The visual encoder extracts image features and the textual decoder generates HTML table tags based on the image features. 

\paragraph{Visual encoder.}
In the existing literature~\cite{nassar2022tableformer, huang2023improving}, the visual encoder is a stack of transformer encoder layers, preceded by a \ac{cnn} backbone. 
In our visual encoder, the \ac{cnn} backbone is supplanted by a linear projection following the design of the \ac{vit}~\cite{dosovitskiy2020image}. 
The linear projection layer reshapes the image $I \in \mathbb{R}^{H \times W \times C}$ into a sequence of flattened 2D patches $I_p \in \mathbb{R}^{N \times (P^2 \cdot C)}$, where $C$ is the number of channels, $(P, P)$ is the size of each image patch, and $N = HW/P^2$ is the number of patches, which is also the input sequence length for the transformer of the textual decoder. 
It is implemented by a stride $P$, kernel $P \times P$ convolution applied to the input image.

\paragraph{Textual decoder.}
The table structure can be defined either using HTML tags~\cite{li2019tablebank} or LaTeX symbols~\cite{ye2021pingan}. 
Since these different representations are convertible, we select HTML tags as they are the most commonly used format in dataset annotations~\cite{zhong2020image, zheng2021global,li2019tablebank}. 
Our HTML structural corpus has 32 vocabularies, including
(1) starting tags \texttt{<thead>}, \texttt{<tbody>}, \texttt{<tr>}, \texttt{<td>}, along with their corresponding closing tags;
(2) spanning tags \texttt{<td}, \texttt{>} with the maximum values 
 for \texttt{rowspan} and \texttt{colspan} set at 10;
(3) special tokens \texttt{<sos>}, \texttt{<eos>}, \texttt{<pad>}, and \texttt{<unk>}. 
The textual decoder is a stack of transformer decoder layers that takes embedded features directly from the visual encoder outputs.  
During training, we apply the teacher forcing so that the transformer decoder receives ground truth tokens. 
At inference time, we employ greedy decoding, using previous predictions as input for the transformer decoder.

\subsection{SSP for the Visual Encoder}
We pre-train the visual encoder with \ac{ssp} and finetune the entire \ac{tsr} transformer model as we find pre-training the visual encoder alone is sufficient to mitigate the performance gap between the hybrid \ac{cnn}-transformer and the linear projection transformer. 
The goal of \ac{ssp} is \ac{mim}, where an image has two views: image patches and visual tokens. 
The image patches are the direct output of the linear projection layer in the visual encoder, and the visual tokens are extracted from the visual codebook of a trained \ac{vqvae}. 
DALL-E~\cite{ramesh2021zero} provided a pre-trained \ac{vqvae} on natural images. 
Given that there is a domain mismatch between natural and tabular images, we cannot directly use the pre-trained weights from DALL-E~\cite{ramesh2021zero} and train a \ac{vqvae} from scratch.
In the sections below, we will first introduce the training of \ac{vqvae}, and then present the pre-training objective of the visual encoder. 

\paragraph{Training \ac{vqvae}}
The \ac{vqvae} architecture consists of two modules: tokenizer and decoder. 
The tokenizer $q_{\phi}(z | I)$ maps the input image $I$ into discrete tokens $z$ in the latent embedding space according to a visual codebook. 
The decoder $p_{\psi}(I | z)$ reconstructs the input image $I$ based on the discrete visual tokens generated by the tokenizer.
The overall reconstruction objective is written as $\mathbb{E}_{z \sim q_{\phi}(z | I)} [ \log p_{\psi}(I | z) ]$.
Similar to the natural language, an image is represented by a sequence of discrete visual tokens $z \in \mathbb{R}^{K \times D}$ in the latent space, where $K$ is the size of the discrete latent space, \ie, a $K$-way categorical, and $D$ is the dimension of each latent embedding vector.
Since the latent representations are discrete, the training process is non-differentiable.
Gumbel-Softmax relaxation is employed in the \ac{vqvae} training, which is a re-parameterization trick for a distribution that smoothly deforms it into the categorical distribution~\cite{jang2016categorical}. 

\paragraph{Pretraining objective of the visual encoder.}
The \ac{ssp} objective of the visual encoder is \ac{mim}. 
The \ac{mim} randomly masks a certain portion of the image patches and the model is required to predict the visual tokens of the masked region based on the unmasked patches. 
The labels of the visual tokens are generated by the tokenizer of the pre-trained \ac{vqvae}.

\subsection{Finetuning TSR Transformer}
During finetuning, we train the whole \ac{tsr} transformer including the visual encoder and the textual decoder. 
Since the visual encoder has been pre-trained by \ac{mim}, we apply two finetuning schedules: 1) freezing the weights in the visual encoder and only training the textual decoder for a few epochs, and 2) directly finetuning all the weights. 

\paragraph{Loss Function} 
We formulate the finetuning loss based on the language modeling task because the HTML table tags are predicted in an autoregressive manner. 
Denote the probability of the $i$th step prediction $p(t_{s_i} | I, t_{s_1: s_{i - 1}}; \theta)$, we directly maximize the correct structure prediction by using the following formulation:
\begin{equation}
    \theta^\ast = \argmax_\theta \sum_{(I, T_s)} \log p(T_s | I; \theta),
\end{equation}
where $\theta$ are the parameters of our model, $I$ is a tabular image, and $T_s$ is the correct structure sequence.
According to language modeling, we apply the chain rule to model the joint probability over a sequence of length $n$ as 
\begin{equation}
    \log p(T_s | I; \theta) = \sum_{i=2}^n \log p(t_{s_i} | I, t_{s_1: s_{i - 1}}; \theta).
\end{equation}
The start token $t_{s_1}$ is a fixed token \texttt{<sos>} in both training and testing. 
\section{Experiments}
\label{sec:exp}

\subsection{Settings}
\paragraph{Architecture.} 
The visual encoder has 12 layers of transformer encoders, and the textual decoder has 4 layers of transformer decoders. 
For the baseline hybrid \ac{cnn}-transformer, we follow the optimal design from Nassar, \etal~\cite{nassar2022tableformer}.
All transformer layers have an input feature size of $d =$ 768, a feed-forward network of 3072, and 12 attention heads. 
The patch size is $P = 16$ in the linear projection. 
The maximum length for the HTML sequence decoder is 512.

\paragraph{Training.} 
All models are trained with the AdamW optimizer~\cite{loshchilov2017decoupled} as transformers are sensitive to the choice of the optimizer. 
We employ a cosine learning rate scheduler with a warmup of 5 epochs and a peak learning rate of $1e-4$ for 15 epochs. 
To prevent overfitting, we set the dropout rate to 0.5.
The input images are resized to $448 \times 448$ in the \ac{ssp} by default~\cite{zhong2020image, nassar2022tableformer}, and normalized using mean and standard deviation. 

\paragraph{Dataset and metric.} 
The \ac{ssp} is only trained on the tabular images of PubTabNet~\cite{zhong2020image} and 
SynthTabNet~\cite{nassar2022tableformer}. 
The finetuning is trained and tested on PubTabNet~\cite{zhong2020image} with HTML annotations. 
The PubTabNet and SynthTabNet are the two largest publicly accessible \ac{tsr} datasets, with $\sim$560k and $\sim$600k image-HTML pairs, separately. 
The evaluation metric is \ac{teds} score~\cite{zhong2020image}.
It converts the HTML tags of a table into a tree structure and measures the edit distance between the prediction $T_{pred}$ and the groundtruth $T_{gt}$:
\begin{equation}
    \text{TEDS}(T_{pred}, T_{gt}) = 1 - \frac{\text{EditDist}(T_{pred}, T_{gt})}{\max (\lvert T_{pred} \rvert, \lvert T_{gt} \rvert)},
\end{equation}
A shorter edit distance indicates a higher degree of similarity, leading to a higher \ac{teds} score. 
Tables are classified as either ``simple'' if they do not contain row spans or column spans, or ``complex'' if they do.
We report the \ac{teds} scores for simple tables, complex tables, and the overall dataset. 

\paragraph{Results}
As shown in Table~\ref{tab:sota}, training the linear projection transformer from scratch leads to a significant performance drop in \ac{teds} score compared to the hybrid \ac{cnn}-transformer architecture, especially for complex tables. 
We present the results of two aforementioned finetuning methods, and both methods achieve similar performance. 
Compared to the \ac{sota} methods, our model outperforms TableFormer and is on par with VAST. 
Clearly, the performance gap between the linear projection transformer and the hybrid \ac{cnn}-transformer is successfully mitigated by our \ac{ssp} method, which demonstrates the effectiveness of \ac{ssp}.

\section{Conclusion}
In this work, we discover that the performance gap between the linear projection transformer and the hybrid \ac{cnn}-transformer can be mitigated by \ac{ssp} of the visual encoder in the \ac{tsr} model. 
We conducted reproducible ablation studies to enhance transparency, inspire innovations, and facilitate fair comparisons in our domain as tables are a promising modality for representation learning.

\clearpage
\bibliography{reference}

\end{document}